\documentclass[10pt]{article}

\usepackage{lineno,hyperref}
\modulolinenumbers[5]

\usepackage{caption}
\usepackage{subcaption}
\usepackage {amsmath,amsthm,amssymb}
\usepackage{algorithm}
\usepackage{algpseudocode}
\usepackage{cleveref}

\usepackage{multirow}
\usepackage {graphicx}

\usepackage{xcolor}
\usepackage{graphicx}

\newtheorem{proposition}{Proposition}

\usepackage [letterpaper]{anysize}

\marginsize{1.1in}{1.1in}{.45in}{0.9in}

\usepackage {times}

\def\argmin{\mathop{\rm arg\,min}}

\def\st{\mbox{s.~t.}}

\def\E{{\mathsf E}}

\def\x{{\bf x}}
\def\y{{\bf y}}
\def\z{{\bf z}}

\def\X{{\bf X}}
\def\Y{{\bf Y}}

\usepackage{color}

\begin{document}









\title{\bf Predictive machine learning for prescriptive applications: a coupled training-validating approach}




\author{
Ebrahim Mortaz\dag\thanks{Corresponding author. E-mail:
emortaz@pace.edu}
\and
Alexander Vinel\ddag
}

\date{\small
\dag  Lubin School of Buisness, Pace University, New York, NY 10038 \\
\ddag Department of Industrial and Systems Engineering, Auburn University, Auburn, AL 36849 
}


\maketitle


\begin{abstract}

In this research we propose a new method for training predictive machine learning models for prescriptive applications. This approach, which we refer to as  coupled validation, is based on tweaking the validation step in the standard training-validating-testing scheme. Specifically, the coupled method considers the prescription loss as the objective for hyper-parameter calibration. This method allows for intelligent introduction of bias in the prediction stage to improve decision making at the prescriptive stage, and is generally applicable to most machine learning methods, including recently proposed hybrid prediction-stochastic-optimization techniques, and can be easily implemented without model-specific mathematical modeling.
Several experiments with synthetic and real data  demonstrate promising results in reducing the prescription costs in both deterministic and stochastic models.



\end{abstract}


\section{Introduction}
\label{intro}

A decision making under uncertainty task usually entails selecting an action (or a set of actions) given that its outcome is dependent on some uncertain events~\cite{kochenderfer}. The relationship between action and the outcome of uncertainty is usually  expressed in the form of historical data and two ways to utilize it can be specified. The first, stochastic optimization, assumes that the uncertainty is modeled as a random vector (with distribution determined based on the historical observations), and then proposes to select decisions than optimize certain statistics of either cost or reward of interest, most often the average ~\cite{ruszczynski}. The second approach, which can be referred to as predictive analytics, purports to forecast the uncertain outcome by training a predictive model on the historical observations and then making a perdition based on some auxiliary observations, and consequently, select the optimal decision for the outcome predicted.

The former approach is widely studied in the operations research and management science communities, and is usually favorably contrasted to the deterministic optimization, where decisions are made without accounting for uncertainty (for example, based on the expected scenario). Naturally, taking uncertainty into account allows for hedging against unfavorable outcomes, leading to improved performance in the long run. On the other hand,  constructing an adequate stochastic model can be very difficult and often requires a significant investment in the historical data. Further, the proposed decisions may turn out to be too conservative, i.e., ill-suited for the actual realized uncertainty outcomes. 

The latter methodology is more common in the  
computer science community, and naturally relies on the ability of the decision maker to train an accurate predictive model. Once the model is constructed, the decisions are usually selected by applying a deterministic model, i.e., the forecast is assumed to be the ground truth. Naturally, it requires some auxiliary information that can then be correlated with the future outcomes to be viable. Such an approach can be expected to outperform classic stochastic optimization if the forecasting accuracy is high enough, i.e., when the decision maker can be sufficiently sure that the predicted decisions are indeed going to be realized, and hence hedging is unnecessary. On the other hand, if accuracy is low, then a more conservative approach based on all of the possible outcomes may be preferred (see some discussion below on hybrid approaches).

These observations naturally explain why forecasting accuracy (along with sensitivity, specificity or their combinations, which we loosely treat as proxies for accuracy for the purposes of this discussion) is the primary consideration in training  predictive models, since it can often directly translate into improved decisions. At the same time, it is also easy to surmise that awareness of the fact that the ultimate goal of the prediction may be  to inform decision making can be beneficial. Consequently, we hypothesize that training framework that accounts for the prescriptive step can result in improved decision making, which serves as the primary  motivation for  our proposed approach, and our goal is to  study approaches for  enabling such training methodology.

  To this end, we propose what we refer to as \emph{coupled validation framework}, whereby prediction accuracy (training error) is optimized at the model training step, while the prescription cost   (i.e., the corresponding decision cost or negative reward) is used at the \emph{validation} step. Recall, that validation (e.g., cross-validation) is commonly used in machine learning modeling as a way to select model hyperparameters, often as a way to avoid over-fitting. Instead, we pose that if the consequent prescription cost structure is available at the time of training, it can be beneficial to use it instead of prediction error as the objective at the validation step, i.e., validation should be designed as the problem of optimizing the prescription cost over the feasible space for the  hyperparameters. This approach is contrasted with the usual validation procedure, referred to here as \emph{decoupled}, which employs prediction error as the objective in hyperparameter tuning step. 

A research direction that is particularly relevant for our effort is a recent approach due to Bertsimas et al.\ \cite{Bertsimas}, where the authors also consider the  apparent disconnect between the deterministic decision making prevalent in machine learning-based decision making and the  uncertain nature of many real-world settings. Specifically, a special machine learning framework is proposed there, which entails a  combination of stochastic optimization and deterministic machine learning-based thinking, whereby machine learning  (and auxiliary observations) is used to derive  the stochastic model of uncertainty, which is then employed in the standard sample average approximation (SAA) algorithm to select a decision that is both hedging against the uncertainty and is informed by the  ML forecasting ability. The approach can be a powerful  tool  for decision making as it allows for incorporating advantages of both frameworks. As we discuss in the consequent sections, our proposed coupled validation approach is directly applicable for this methodology as well. Namely, the validation based on prescription cost is naturally possible in the case of ML-SAA approach due to Bertsimas et al. in ~\cite{Bertsimas}, and can lead to improved performance.

The main contribution of this effort is in proposing the coupled validation framework for training machine learning models that will be used in optimization. The framework is applicable to most common machine learning models or even hybrid machine learning and stochastic optimization approaches. Further, it does not rely on model-specific knowledge and can be automated. 
The rest of manuscript is organized as follows. In the remainder of \textbf{Section 1}, we discuss a motivating example and the relevant literature. In \textbf{Section 2} we offer the proposed methodology that includes problem statement and theoretical analysis of the coupled approach. \textbf{Section 3} presents the results of experiments with synthetic and real detests and finally \textbf{Section 4} provides concluding remarks and a discussion of study limitations.

\subsection{Motivating example}

To illustrate the proposed approach, we next consider a motivating example. Consider a classic newsvendor problem, for example as presented in~\cite{ban}. Specifically, assume that the newsvendor is facing a random demand value $Y$ with the purchasing quantity decision $z$ that should be selected a day ahead before demand realization and given pre-specified shortage/surplus cost.
In this classic setting the problem has a well-defined analytic solution if the distribution of $Y$ is given.
Now, further assume that it is possible for the newsvendor to access at the time of decision making auxiliary information $x$ that is correlated with demand. For example, $x$ can represent weather forecast, day of the week, etc. In this case, it may be possible to construct a machine learning (or statistical) predictive model $\hat{y}(x)$ for forecasting the demand. Note though, that while naturally a more accurate prediction can directly translate to better purchasing decision, there exists an asymmetry in the cost of under- or over-predicting the demand. Since in newsvendor problems the surplus cost (wasted inventory) is usually higher than the shortage cost (lost potential revenue), in this case it is natural to design the predictive model to penalise the corresponding prediction error higher. Naturally, it is possible to manually organize training in such a way as to enable this by design by introducing an appropriate penalty parameter. In this case, a predictive model that is biased towards underestimating demand may lead to  improved decision making. 

In general a careful design of penalties (or other training procedures) to enable this effect may challenging. Indeed, in general, the relationship between the  prediction, decisions and costs can be complicated and harder to understand. Consequently, a general approach for incorporating prescriptive information into training may be beneficial.
To this end, we consider the coupled learning framework aimed at proposing such an approach. The formal presentation is given in Section \ref{sec:methodology}. As a preliminary discussion of the approach we summarize and contrast the coupled and decoupled methods in Table~\ref{tab:Dsteps}, which describes the two approaches within the usual training-testing-validation framework. Recall, that usually the available historical data is separated into training, validation and testing sets. The training set is used to determine the model parameters, with the goal of reducing prediction error. Validation set is then used to select hyper-parameters for the model selected. Finally, the testing set is used to evaluate the resulting prediction quality. 
In the proposed framework, we surmise that the 
validation step is particularly suitable for accounting for the optimization aspect. 
Indeed, regardless of the prescriptive goal, it is  easy to observe that prediction accuracy should remain the main goal of the machine learning step, since accurate prediction is the main factor that enables the improvement in prescriptive actions compared to simple stochastic approach. Consequently, we conclude that if the prescriptive cost is used as the objective in the   validation step, both criteria (accounting for prescription cost and prediction accuracy) can be fulfilled at the same time.

\begin{table}[h!]
    \small
\begin{center}
	\caption{Training, validating and testing steps for the coupled/decoupled approaches in a deterministic prediction model}
	\label{tab:Dsteps}
\resizebox{0.9\textwidth}{!} {%
	\begin{tabular}{l|l}
		\hline
Coupled& Decoupled\\
\hline
\multicolumn{2}{l}{\textbf{1.} Divide the dataset into a training, a validation and a testing sets randomly.}\\
\multicolumn{2}{l}{\textbf{2.} Train the \textit{predictive} models using random hyperparameteres and the training set.}\\
\textbf{3c.} Find the hyperparameteres that give minimum &\textbf{3d.} Find the hyperparameteres that give minimum   \\
\quad \quad  \textit{prescriptive} loss using the validation set.& \quad \quad  \textit{predictive} loss and the validation set.\\
\multicolumn{2}{l}{\textbf{4.} Test the prescriptive loss using the parameters from steps \textbf{3c}, \textbf{3d} and the testing set.}\\
	\end{tabular}}
	\end{center}
\end{table}

\subsection{Relevant literature}

Machine learning research  has been attracting a significant amount of attention with  applications spanning from predicting to grouping, profiling, matching, clustering, data reduction and  many others~\cite{jordan}. The predictive ML models are specifically broken down into two major categories: supervised and unsupervised. In the supervised models, learning is guided by an outcome variable/label~\cite{friedman}. The supervised models are similarly divided into two major subcategories: regression and classification. When the outcome variable of a supervised ML is quantitative, the problem is a regression~\cite{tibshirani}. In training regression models, the goal is usually in finding optimal estimators that minimize the average loss/error, defined as the difference between the actual and predicted outcomes~\cite{lehmann}. Such ML methods could include ordinary linear regression, ridge regression~\cite{hoer}, the lasso~\cite{tibshirani} or more advanced regressors such as artificial neural networks. There also exists a non-parametric class of regression models that are trained to find a neighborhood of similar observations to the observation in hand for prediction. These models include supervised clustering models such as $k$NN~\cite{altman} or tree-based models such as regression trees \cite{breiman}, bagged trees~\cite{breiman_bagging}, random forests~\cite{breiman_random} and boosted trees~\cite{friedman_greedy}.

On the other hand, stochastic programming is an advanced tool to model decision making under uncertainty~\cite{birge}, with significant amount of research dedicated to developing better techniques to solve such problems. Replacing the true (usually unknown) distribution of the random parameters with a set of historical~\textit{iid} observations, a mechanism referred to as the SAA method~\cite{kleywegt}, is one of the most well-established methods that allows the decision maker to solve a stochastic program with the off-the-shelf software.  In this paper, we only focus on the two-stage stochastic programs~\cite{shapiro} and the SAA method when discussing stochastic optimization problems, not considering alternative frameworks, such as  stochastic dynamic programming \cite{bertsekas} for multi period decision making, or robust optimization~\cite{nemirovski}.
In stage one, the ‘here and now’ decisions are made and are then followed by the stage two decisions which are ‘wait and see’ or recourse actions~\cite{shapiro}. However, as argued in the previous section, even well-developed stochastic methods with SAA can be computationally demanding~\cite{shapirocomplexity} and rely on the extensive historical data to adequately describe the random distributions. In the absence of such data, the resulting decisions may be too conservative. Consequently, in many practical applications the stochastic modeling methodology is abandoned in favor of simpler  deterministic programs~\cite{murty}.

Naturally, if it is possible to use some additional information available at the time of decision making to predict the realizations of uncertain parameters (using machine learning or statistics), then the deterministic decision making quality can be significantly improved.  
At the same time, there is no well-established methodology in the literature that outlines approaches and best practices to using ML models for prescription tasks. In the context of deterministic programming, training the ML models for prescription tasks are often carried out in sequence, usually referred to as the ``predict-then-optimize'' scheme. There is a rich literature discussing the application of this scheme in different fields of business and engineering. For instance, in~\cite{chen}, an ML based method is proposed for a system dynamics problem in which the time-varying parameters are unknown and need to be predicted. Using the historical system data and ML models, the parameters are estimated and then are fed into a model predictive control to optimize some performance indicators.  Another engineering application of this scheme is provided by Xu et al. in~\cite{xu} where an ML-based wear fault diagnostic model is used to predict fault modes in marine diesel engines, which then enables taking quick actions to avoid severe accidents in ships. An example of business application is given in ~\cite{churn}, where the authors introduced a new step-wise regression methodology to better manage a company's customer churn and improve decision making in marketing to help customer retention. Note though that in most of such cases, the training and prescription steps are taking consequently without informing each other.

Within the context of stochastic programming, the traditional methods are centered on how to use historical observations of uncertain parameters) and paid little attention to any possible auxiliaries that may impact the uncertainty. Kao et al. (2009) \cite{kao} proposed a method to train an ML model that minimizes loss with respect to a nominal optimization problem, specifically, an unconstrained quadratic optimization program. Later, Ban et al. in~\cite{ban}, introduced a mathematical method that trains ML models to directly predict the optimal solution of a newsvendor problem from big data, again with an unconstrained optimization problem. These limiting assumptions on the optimization problems sparked interest in other researchers to seek a more general approach. In \cite{tulabandhula}, Tulabandhula et al. propose to minimize operational cost and prediction error of the model with constraints. However, the operational costs are computed based on the predicted and not the true parameters. Later, Bertismas et al. \cite{Bertsimas} proposed a new weight structure for the sample data derived from training several ML models (some parametric, some non-parametric) in a stochastic optimization model. Their offered framework is similar to ``predict and then optimize'' in spirit, however, unlike that scheme, where the historical ${\Y}$ values are equiprobable,  in the proposed structure they are weighted. A drawback for the proposed framework is that the weight values cannot be derived from all types of ML models. One of important class that does not appear to fit that structure is the artificial neural network. The other issue is that the prescriptive loss has no bearing in predictive learning. Motivated by these results  Elmachtoub et al. in ~\cite{elmachtoub} proposed a new framework, called ``smart predict, then optimize'', refereed to as SPO. The SPO leverages the optimization problem in the prediction learning via a loss function that measures the decision error induced by the prediction. The offered framework is similar to relative regret concept introduced in~\cite{lim}. One issue with the approach is that the unknown parameters needed to appear linearly in the objective function of a stochastic optimization problem as the solution technique needed to create a convex surrogate to solve the problem with duality theory. An extension of the proposed methodology called SPO Trees (SPOTs) for training decision trees under this loss is offered again by Elmachtoub et al. in ~\cite{elmachtoub2}.  Another extension to the smart SPO approach is presented by Mandi et al. in ~\cite{mandi} where the use of SPO to solve a relaxed version of a combinatorial optimization problems is investigated. An application of the smart SPO denoted as ``semi SPO'' is proposed by \cite{yan} for efficient inspection of ships at ports in maritime transportation. A loss function motivated by the proposed approach in~\cite{elmachtoub} is utilized to predict the number of deficiencies each inspector can identify for each ship. 

We argue that one drawback of the smart SPO and its extensions as well as the hybrid predictive prescriptive framework proposed by Bertsimas et al. in~\cite{Bertsimas} is in requiring additional problem-specific modeling. Secondly, the modeling approaches rely on  certain model properties, e.g., convexity, and/or certain ML model structures, for instance, non-parametric ML models. Unlike these methods, the coupled method proposed here, does not require any additional modeling and is meant to be used as a natural transition from training generic machine learning models for predictive to training machine learning models for prescriptive applications. This method is largely designed to be simple and straightforward and to be used by both management science and computer science researchers and practitioners.

\section{Methodology and theoretical  analysis}\label{sec:methodology}

We will use the following notation throughout the rest of the manuscript. Note that we use bold symbols to refer to vectors and capital non-italicised letters for random quantities. 
\begin{itemize}
	\item $\z$ is the decision vector;
	\item $\Y$ is a random vector for the model parameters (e.g., demand), that cannot be observed before making decision;
	\item $\X$ is a random vector of auxiliary features that can be observed in advance;
	\item $g(\z, \y)$ is the cost of action $\z$ given realization $\y$ of the random vector $\Y$;
	\item $\mathcal{Z}$ is the feasible region for the decision  $\z$
	\item $(\X^r, \Y^r), (\X^v, \Y^v), (\X^t, \Y^t)$ are random vectors for the training, validation and testing values of the auxiliaries and and model parameters;
	\item $\hat{\y}$ is the ML prediction for vector $\Y$.
\end{itemize}



The following decision problem is considered
\begin{align}
\zeta^*(\Y) = \min_{\z \in \mathcal{Z}} \quad & g(\z, \Y) 
\end{align}
and accordingly
$\z^*(\Y) = \argmin_{\z \in \mathcal{Z}} \Big\{  g(\z, \Y) \Big \}$.
Note that, as formulated, the problem is deterministic for a given realization of $\Y$.
In a stochastic setting, decision maker's goal is to select action $\z$ given the uncertainty in the realization of $\Y$. The usual stochastic optimization approach is to formulate the expected cost minimization problem as
\begin{align}
\zeta_{SAA}^* = \min_{\z \in \mathcal{Z}} \quad & \E_\Y \Big [ g(\z, \Y)\Big ].
\end{align}
Note that here $\z$ is treated as the first-stage decision and we omit any potential recourse actions, which can be added to the problem formulation without impacting the rest of the discussion. 
Alternatively, a perfect information solution can be constructed as 
\begin{align}
\zeta_{PI}^* = \E_\Y \Big [\min_{\z\in \mathcal{Z}} \quad &  g(\z, \Y)\Big ].
\end{align}
Naturally, it is easy to see that $\zeta_{PI}^* \le \zeta_{SAA}^*$.

\paragraph{Predictive model construction}

Now, suppose that the decision maker is able to obtain a realization of random vector of auxiliary features $\X$ before action $\z$ needs to be taken. Assuming that realization of $\X$ is correlated with realization of $\Y$, a machine learning model can be constructed, denoted as
$
\hat{\y}(\X, \beta, \gamma) = \hat{f}(\X ,\beta, \gamma),
$
where $\beta$ are the ML model parameters selected based on the training data, and $\gamma$ are the hyperparameters selected through validation. 
Specifically, if we define prediction loss function as 
$
L(\Y, \hat{\y} ) = || \Y - \hat{\y}  ||_2^2
$
then parameters $\beta$ are usually selected as
\begin{align*}
\hat{\beta}(\gamma) = \argmin_\beta \E_{\X^r, \Y^r} \bigg[ L\Big ( \Y^r, \hat{\y}(\X^r, \beta, \gamma) \Big ) \bigg ],
\end{align*}
where $\X^t$ and $\Y^t$ are the random variables representing the features and outcome drawn from the training set distribution.

\paragraph{Optimization with the predictive model}

The decision maker can then employ the machine learning model to obtain prediction $\hat{\y}$ based on the specific realization of auxiliary $\X$, which is then assumed to represent the best guess on  the ground truth. 
Consequently the following problem is solved for determining the best course of actions.
\begin{align}
\zeta^*(\X, \beta, \gamma) =  \min \quad & g\left(\z,\hat{\y}(\X|\beta, \gamma)\right) \\
\st \quad & \z \in \mathcal{Z}\\
&\hat{\y}(\X, \beta, \gamma) = \hat{f}(\X,  \beta, \gamma).
\end{align}
and accordingly
\begin{align}
\z^*(\X, \beta, \gamma) =  \argmin \quad & g\left(\z,\hat{\y}(\X|\beta, \gamma)\right) \\
\st \quad & \z \in \mathcal{Z}\\
&\hat{\y}(\X, \beta, \gamma) = \hat{f}(\X , \beta, \gamma).
\end{align}
The corresponding cost  can be calculated as $\zeta^*_{Pred} = \E_\X \Big [ \zeta^*(\X, \beta, \gamma) \Big ]$. Naturally, as before,  $\zeta_{PI}^* \le \zeta_{Pred}^*$, representing the observation that any inaccuracy in prediction directly results in sub-optimal decisions. On the other hand, relationship between  $\zeta_{Pred}^*$ and $\zeta_{SAA}^*$ depends on the accuracy of the ML model.

The approach proposed here is aimed to fine tune the machine learning model to take advantage of the fact that it will be used for prescription.

\paragraph{Two approaches to validation}

Validation in machine learning is usually conducted by selecting values for hyperparameters $\gamma$, which minimize the prediction loss function over the validation set. Note that usually this process is not a comprehensive optimization procedure, and instead, a simpler algorithm is used, for example, random search, where a collection of candidate values of $\gamma$ are tested.  Regardless, we will denote the set of feasible hyperparameters values as $\Gamma$ (a finite set in the case of random search).

\emph{Decoupled validation} represents this traditional approach
\begin{align}\label{eq:decoupledValidation}
\hat{\gamma}^d = \argmin_{\gamma\in \Gamma} \E_{\X^v, \Y^v}\bigg [ L\bigg (\Y^v, \hat{\y} \Big( \X^v,   \hat{\beta}(\gamma), \gamma  \Big ) \bigg ] ,
\end{align}
where $\x^v$ and $y^v$ are the random variables representing the features and outcome drawn from the validation set distribution.

\emph{Coupled validation} is an alternative approach, whereupon the prescriptive loss is used to obtain hyperparameters.
\begin{align}\label{eq:coupledValidation}
\hat{\gamma}^c = \argmin_{\gamma\in \Gamma} \Bigg ( \E_{X^v}\bigg [ \zeta^*\Big (\X^v, \hat{\beta}(\gamma), \gamma\Big ) \bigg ] \Bigg ).
\end{align}
The following two results can be shown to hold. Both follow directly from the definitions and the assumptions.

\begin{proposition}
Coupled validation results in the prescriptive losses over the validation set that do not exceed the losses due to decisions made with decoupled validation, i.e.,
\begin{align}
\E_{\X^v} \Big[
\zeta^*(\X^v, \hat{\beta}(\hat{\gamma}^c), \hat{\gamma}^c)   
\Big ]
\le
\E_{\X^v} \Big[
\zeta^*(\X^v, \hat{\beta}(\hat{\gamma}^d), \hat{\gamma}^d)    
\Big ].
\end{align}

\end{proposition}

\begin{proposition}
Assuming that the distributions of the testing and validation sets are the same $(\X^v, \Y^v) \sim (\X^t, \Y^t)$, coupled validation results in the prescriptive losses over the testing set that do not exceed the losses due to decisions made with decoupled validation, i.e.,
\begin{align}
\E_{\X^t} \Big[
\zeta^*(\X^t, \hat{\beta}(\hat{\gamma}^c), \hat{\gamma}^c)   
\Big ]
\le
\E_{\X^t} \Big[
\zeta^*(\X^t, \hat{\beta}(\hat{\gamma}^d), \hat{\gamma}^d)    
\Big ].
\end{align}
\end{proposition}

Naturally, the law of large numbers implies that the results above also hold for  sampled testing and validation sets, as long as they are
 sufficiently large. These then imply that in theory, in the presence of prescription step, coupled validation approach outperforms the usual decoupled case in terms of the resulting prescription cost. Next we demonstrate that the same conclusions apply to the more advanced hybrid machine learning and stochastic optimization framework due to \cite{Bertsimas}.


Observe that the optimization with prediction approach described above disregards the intrinsic uncertainty in the relationship between $\X$ and $\Y$, i.e., it assumes that the prediction of the machine learning model is the ground truth. In many practical applications this may be an appropriate assumption, especially if the machine learning model  can be trained to a achieve sufficient accuracy, i.e., the auxiliary features are highly correlated with the outcome vector. On the other hand, in many applications, it may be beneficial to design the decision  making problem to hedge against the intrinsic uncertainty stemming from the fact that the auxiliary features may be incapable of capturing the full information on the outcome of $\Y$. SAA-based stochastic optimization problem, described above, is the most conservative approach of this kind, as  it completely disregards $\X$. 
Bertsimas et al. in ~\cite{Bertsimas} describe a middle-ground approach, whereby the machine learning  model is expected to result in a sample (or a distribution) of predictions, which consequently are used withing an SAA approach. 

To this end, we assume that a machine learning model is available, such that its prediction can be interpreted as a random variable with the distribution dependent on the auxiliary features $\X$ as well as parameters $\beta, \gamma$ as before. 
\begin{align}
    \hat{\Y}(\X, \beta, \gamma) = F(\X, \beta, \gamma)
\end{align}
Naturally, optimal solution, that accounts for the uncertainty in the realization of $\hat{\Y}$ can be constructed as
\begin{align}
\zeta_{SAAML}^*(\X, {\beta}, {\gamma}) = \min_{\z \in \mathcal{Z}} \quad & \E_{\hat{\Y}} \Big [ g(\z, \hat{\Y}(\X, {\beta}, {\gamma}))\Big ],
\end{align}
where $\hat{\gamma}$ can be constructed through \emph{decoupled} or \emph{coupled} validation analogously to the discussion above.
Specifically, since $\E[\hat{\Y}(\X, \beta,\gamma)]$ is the best available prediction due to the ML model, as before, $\hat{\beta}(\gamma)$ can be constructed by minimizing the average prediction error between the realization of $\Y^r$ and $E[\hat{\Y}(\X, \beta,\gamma)]$, i.e., 
\begin{align*}
\hat{\beta}(\gamma) = \argmin_\beta \E_{\X^r, \Y^r} 
\Bigg[
L \bigg ( \Y^r, \E_{\hat{\Y}}\Big [\hat{\Y}(\X^r, \beta,\gamma) \Big ]  \bigg )
\Bigg ].
\end{align*}

\emph{Decoupled}  validation then can be defined as 
\begin{align}
\hat{\gamma}^d = \argmin_{\gamma\in \Gamma} \E_{\X^v, \Y^v}\Bigg [ 
L \bigg ( \Y^r, \E_{\hat{\Y}}\Big [\hat{\Y}(\X^r, \hat{\beta}(\gamma),\gamma)\Big ]  \bigg )
\Bigg ].
\end{align}

\emph{Coupled} validation in this case can be defined as 
\begin{align}\label{eq:coupledValidationSAA}
\hat{\gamma}^c = \argmin_{\gamma\in \Gamma} \E_{\X^v} \Big [   \zeta_{SAA-ML}^*(\X^v, \hat{\beta}(\gamma), \gamma) \Big ].
\end{align}
Analogously to the claims above we can conclude that the following result holds, i.e., for the hybrid approach the coupled validation possesses the same properties. 

\begin{proposition}
Assuming that the distributions of the testing and validation sets are the same $(\X^v, \Y^v) \sim (\X^t, \Y^t)$
\begin{align}
 \E_{\X^t} \Big [   \zeta_{SAA-ML}^*(\X^t, \hat{\beta}(\hat{\gamma}^c), \hat{\gamma}^c) \Big ]   \le \E_{\X^t} \Big [   \zeta_{SAA-ML}^*(\X^t, \hat{\beta}(\hat{\gamma}^d), \hat{\gamma}^d) \Big ].
\end{align}

\end{proposition}

\section{Experiments}
For the computational experiments we use two sets of data: synthetic and real. In order to demonstrate that the proposed approach is applicable to a wide variety of ML models, we use the following predictive models: lasso, random forest, gradient boosting machine and fully-connected neural network. Similarly, in order to test that the coupled approach is compatible with the hybrid prediction-optimization methodology due to \cite{Bertsimas}, we employ the \textit{k}NN model following the discussion presented there. In this case, the hyperparameter to be selected during validation corresponds to the number of neighbors $k$ and naturally reflects  the trade off between a purely stochastic model (if all testing data points are included) and a purely predictive model (single closest neighbor).

As a result, we compare the following six decision making approaches: two  purely predictive (deterministic) models based on either coupled or decoupled validation (referred to as D-ML-Coupled and D-ML-Decoupled), coupled and decoupled versions of the hybrid prediction-optimization models (referred to as SAA-ML-Coupled and SAA-ML-Decoupled), as well as standard naive SAA (based on assuming equiprobable scenarios derived from the test set) and perfect information solution. Naturally, the perfect information solution is guaranteed to result in the lowest prescriptive cost by definition and so serves as the baseline for comparing the rest of the solutions. For the sake of completeness, in Tables~\ref{tab:SAA-ML-C} and~\ref{tab:SAA-ML_D} in the Appendix we also present the implementation of the hybrid approach for coupled and decoupled validation schemes. Gurobi~\cite{gurobi} was used to solve all optimization problems  and Scikit-learn~\cite{scikit-learn} for all ML models.


Throughout the section we use classic newsvendor problem as the basis for the experiments, formulated as follows
$
\zeta^*(y) = \min_{z\ge 0} \Big\{ C z + C_{sp}[y-z]_+ + C_{sh}[z-y]_+ \Big\},
%
$
%
%
%
where  $C$, $C_{sh}$ and $C_{sp}$ are the unit ordering and unit shortage, and unit surplus costs respectively, and $[t]_+ = \max\{ 0, t\}$. Note that as formulated the problem is deterministic for a given demand $y$ and can be converted to one of the stochastic formulations as discussed in the previous section. Parameters  $C$, \mbox{$C_{sp}$}, \mbox{$C_{sh}$} are set to 20, 500 and 50 respectively to reflect the significant difference between the shortage and surplus costs.   
All results reported are averaged over ten repeats of 5-fold cross-validation. In all cases we compare the performance of methods based on the average prescription loss on the testing set.

\subsection{Synthetic data}

Synthetic data was generated as $Y = c + f\Big (\sum_i^n x_i \Big) + \text{Uniform}(-a,a)$, where each $x_i$ is generated as a random independent identically distributed sample from the same distribution, $n$ controls the number of predictors and function $f$ is the exact relationship between the predictor and demand, and we use three versions: \textit{linear},  \textit{sinusoidal} and \textit{piece-wise linear} functions.. In other words, the demand $y$ is calculated as a function of the summation of the auxiliaries with random uniform noise.  The relative influence of noise is dependent on the value of parameter $a$, and in the end of this section we evaluate its effect on the performance of the methods.

\begin{figure}[h!]
\centering

           \begin{subfigure}[h]{0.45\textwidth}
        \includegraphics[width=\textwidth]{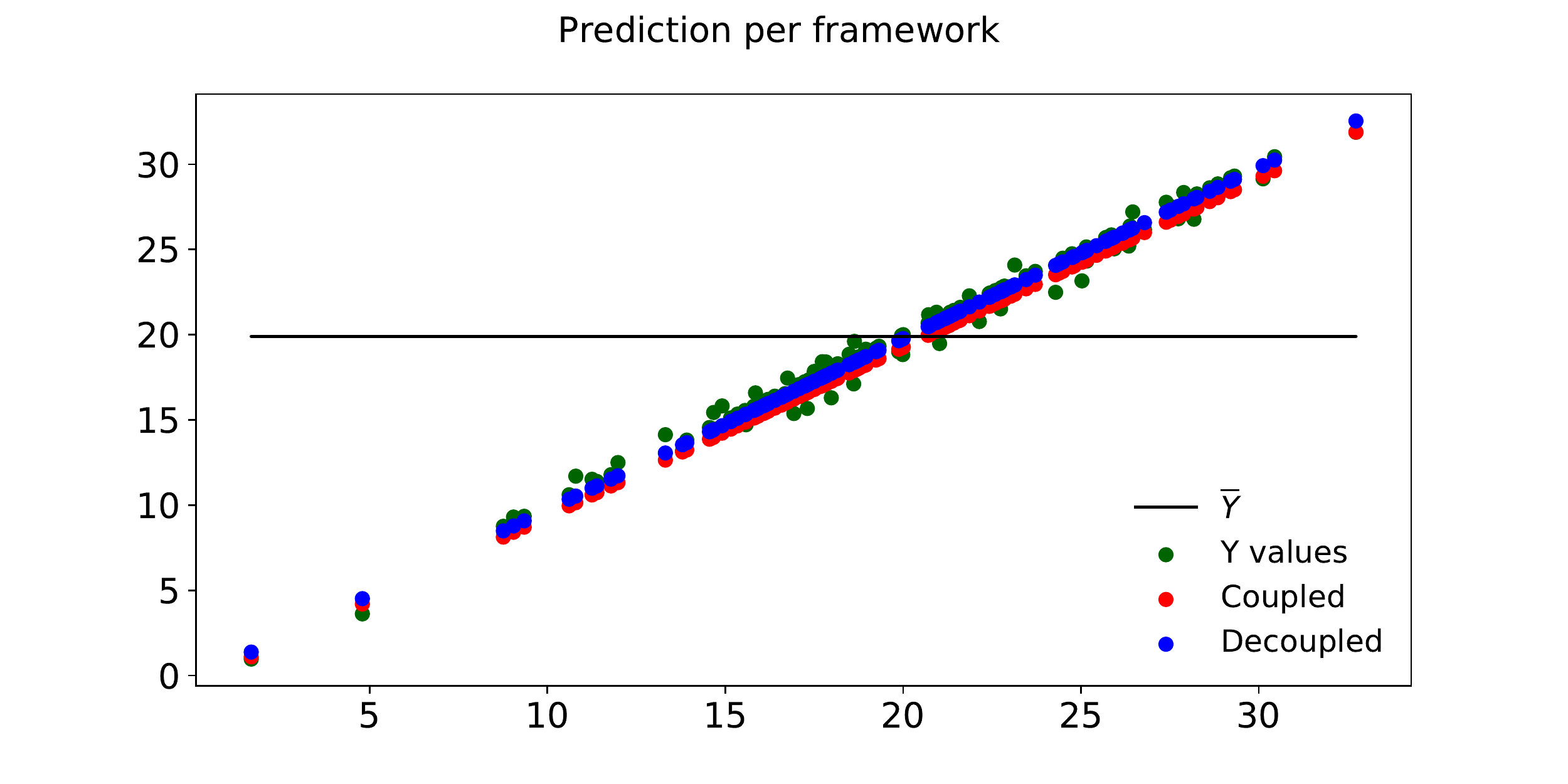}
        \caption{ The Lasso model}
        \label{fig:identity_lasso}       
    \end{subfigure}
    \begin{subfigure}[h]{0.45\textwidth}
        \includegraphics[width=\textwidth]{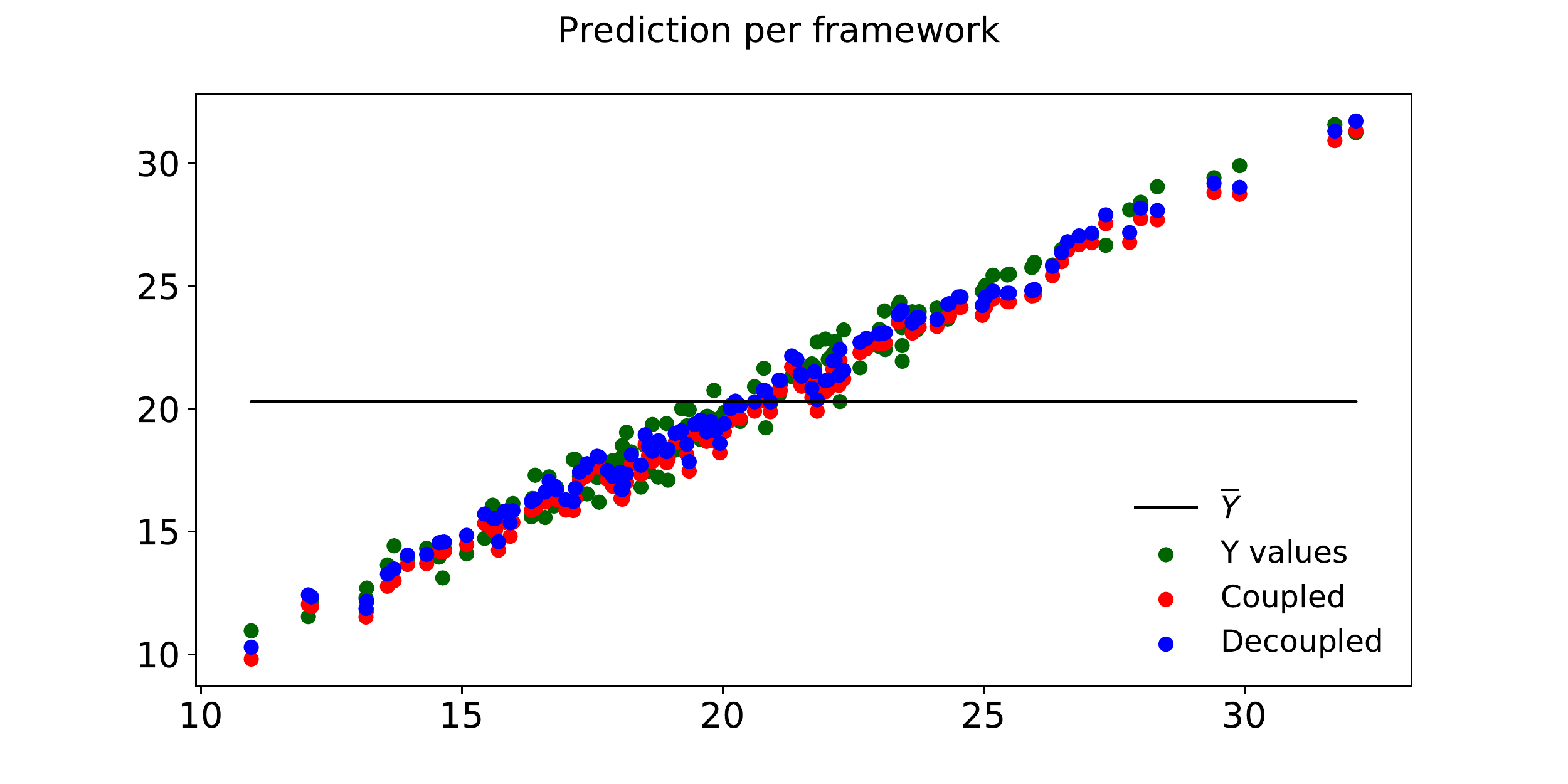}
        \caption{The RF model}
        \label{fig:identity_RF} 
    \end{subfigure}
        \begin{subfigure}[h]{0.45\textwidth}
        \includegraphics[width=\textwidth]{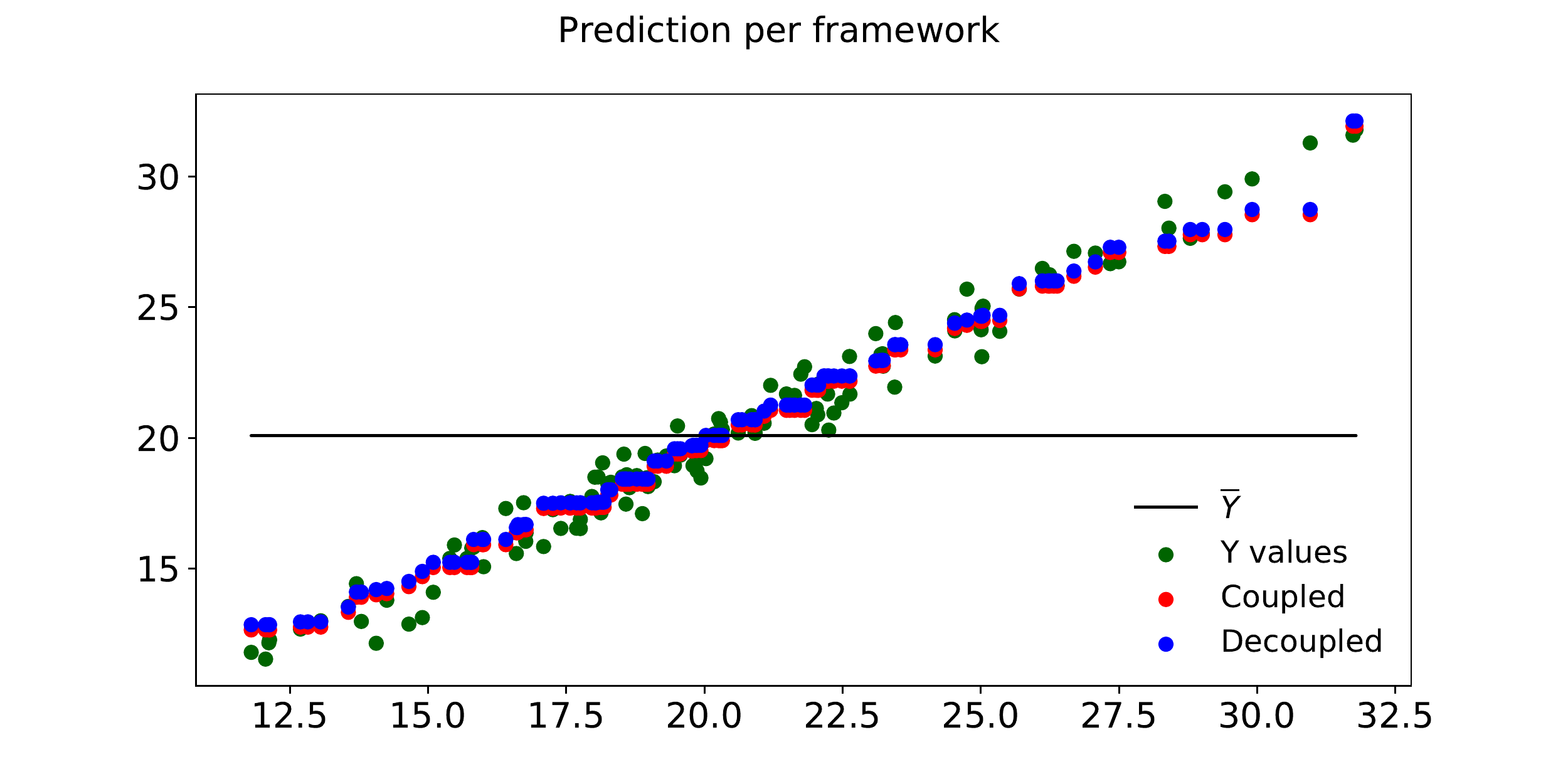}
        \caption{The GBM model}
        \label{fig:identity_GB} 
    \end{subfigure}
    \begin{subfigure}[h]{0.45\textwidth}
        \includegraphics[width=\textwidth]{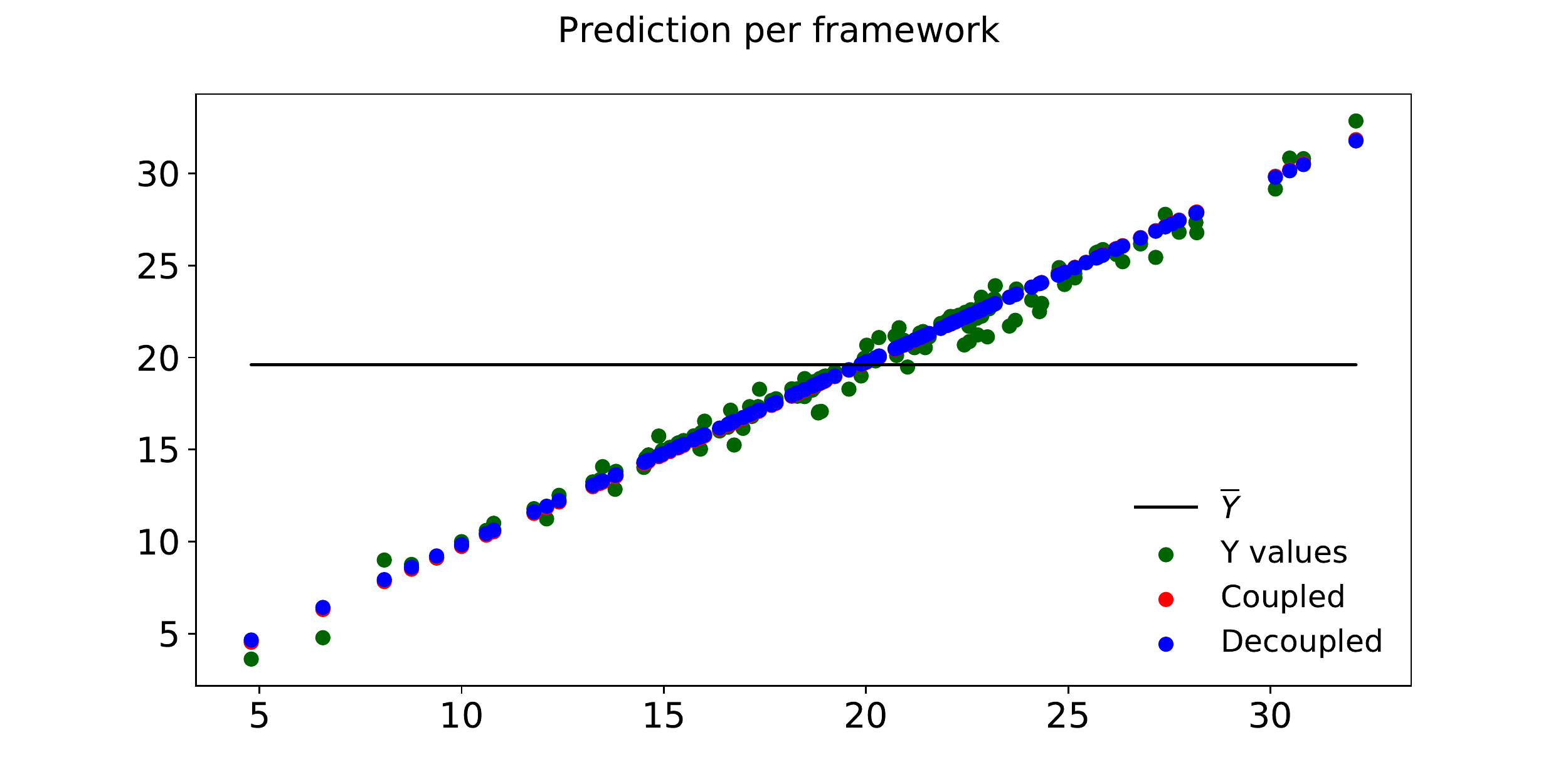}
        \caption{The NN model}
        \label{fig:identity_NN} 
        
    \end{subfigure}
    \caption{Generated data and ML predictions due to coupled and decoupled approaches for the linear sample.}
    \label{fig:identity}
\end{figure}
\begin{figure}[h!]
\centering
\begin{subfigure}[h]{0.45\textwidth}
        \includegraphics[width=\textwidth]{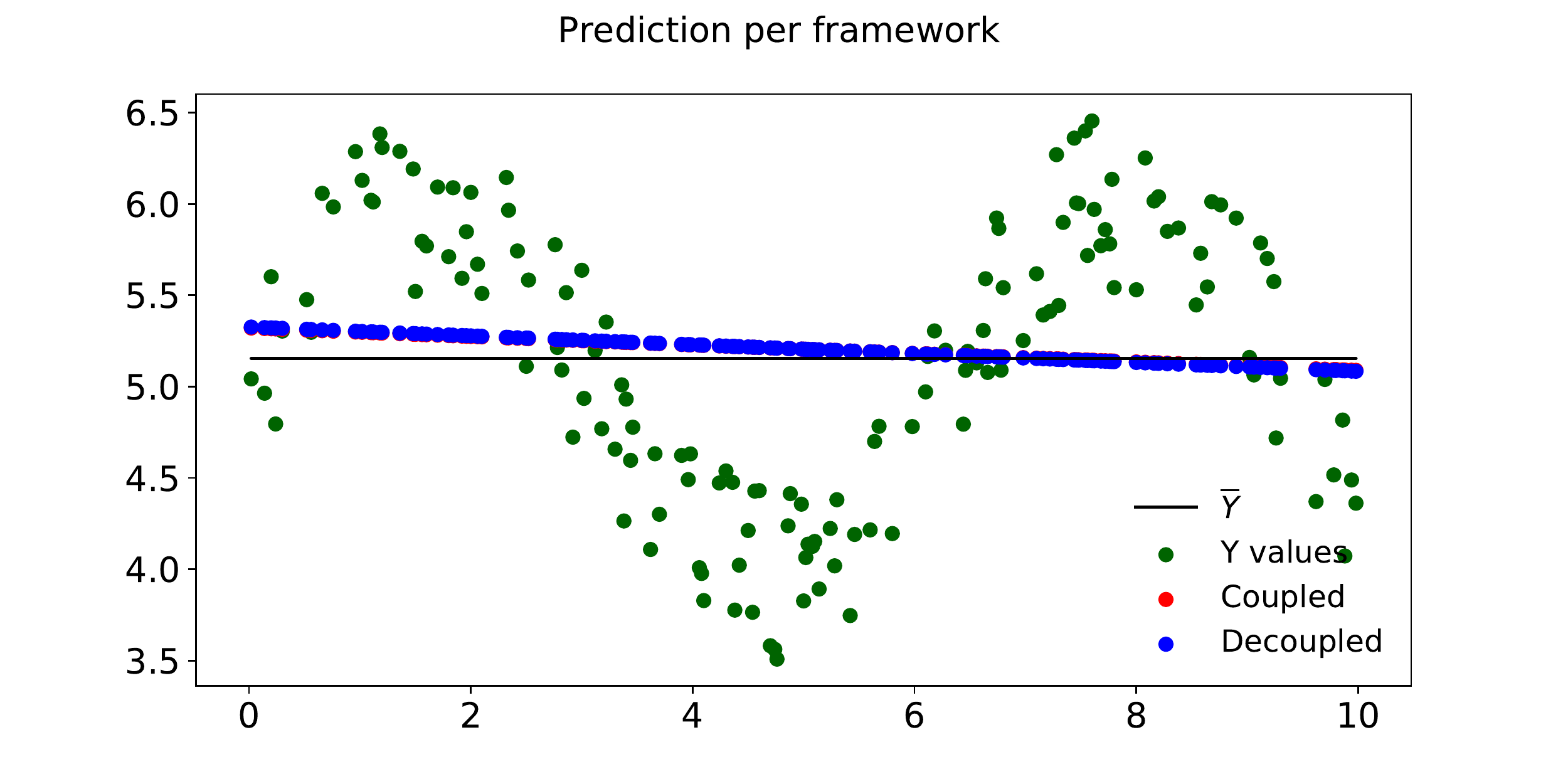}
        \caption{ The Lasso model}
        \label{fig:sin_lasso}       
    \end{subfigure}
    \begin{subfigure}[h]{0.45\textwidth}
        \includegraphics[width=\textwidth]{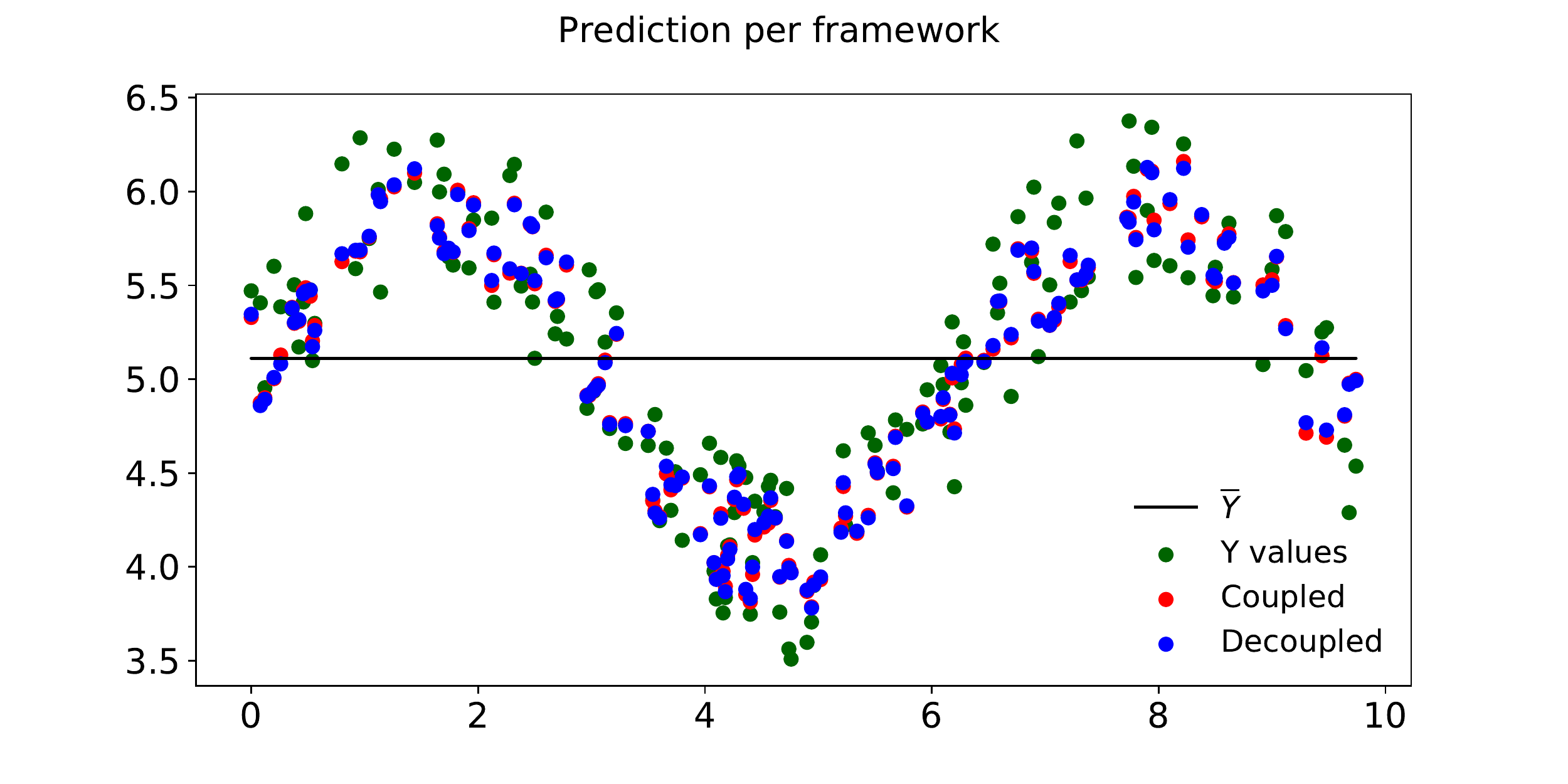}
        \caption{The RF model}
        \label{fig:sin_RF} 
    \end{subfigure}
        \begin{subfigure}[h]{0.45\textwidth}
        \includegraphics[width=\textwidth]{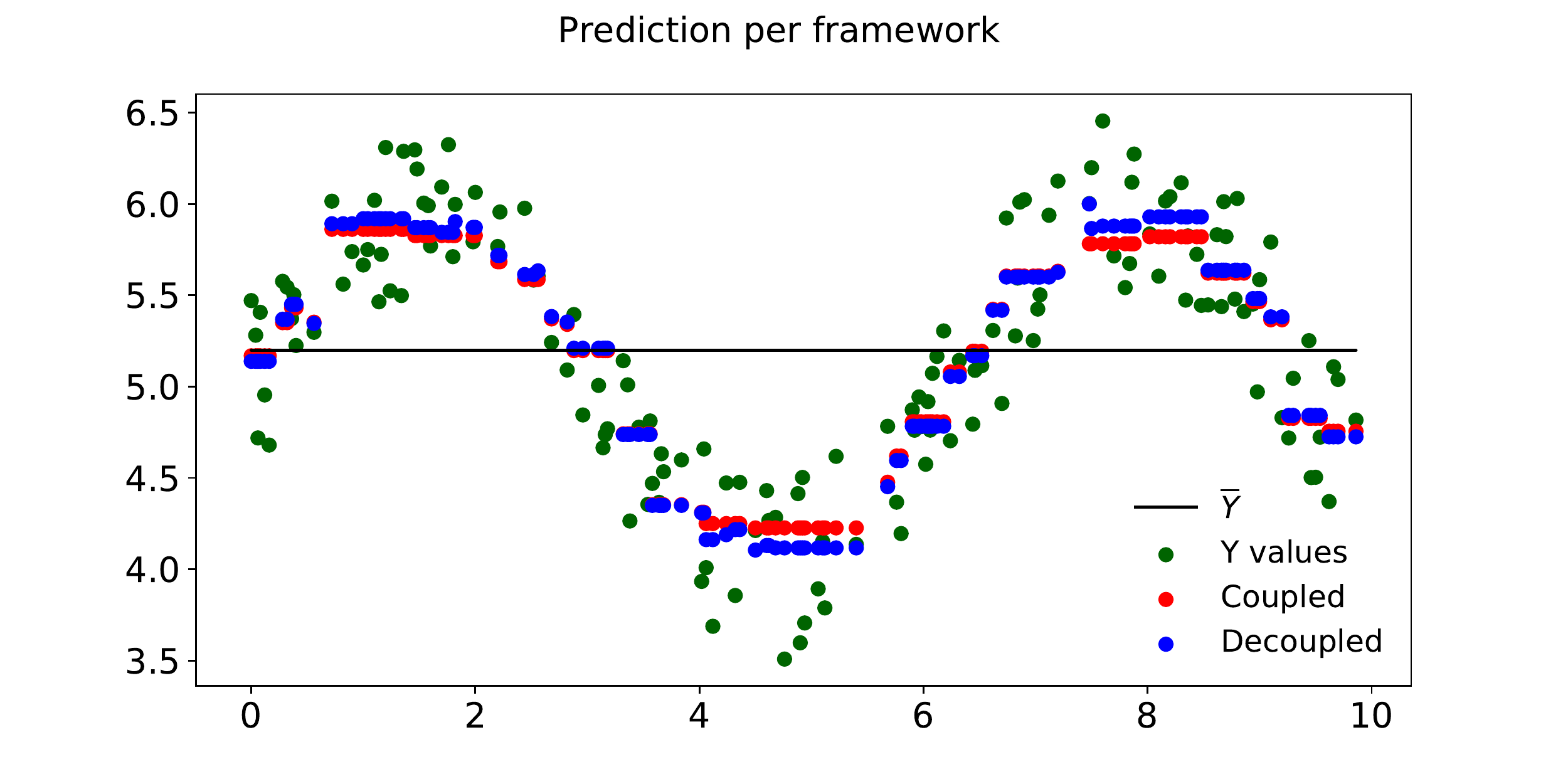}
        \caption{The GBM model}
        \label{fig:sin_GB} 
    \end{subfigure}
    \begin{subfigure}[h]{0.45\textwidth}
        \includegraphics[width=\textwidth]{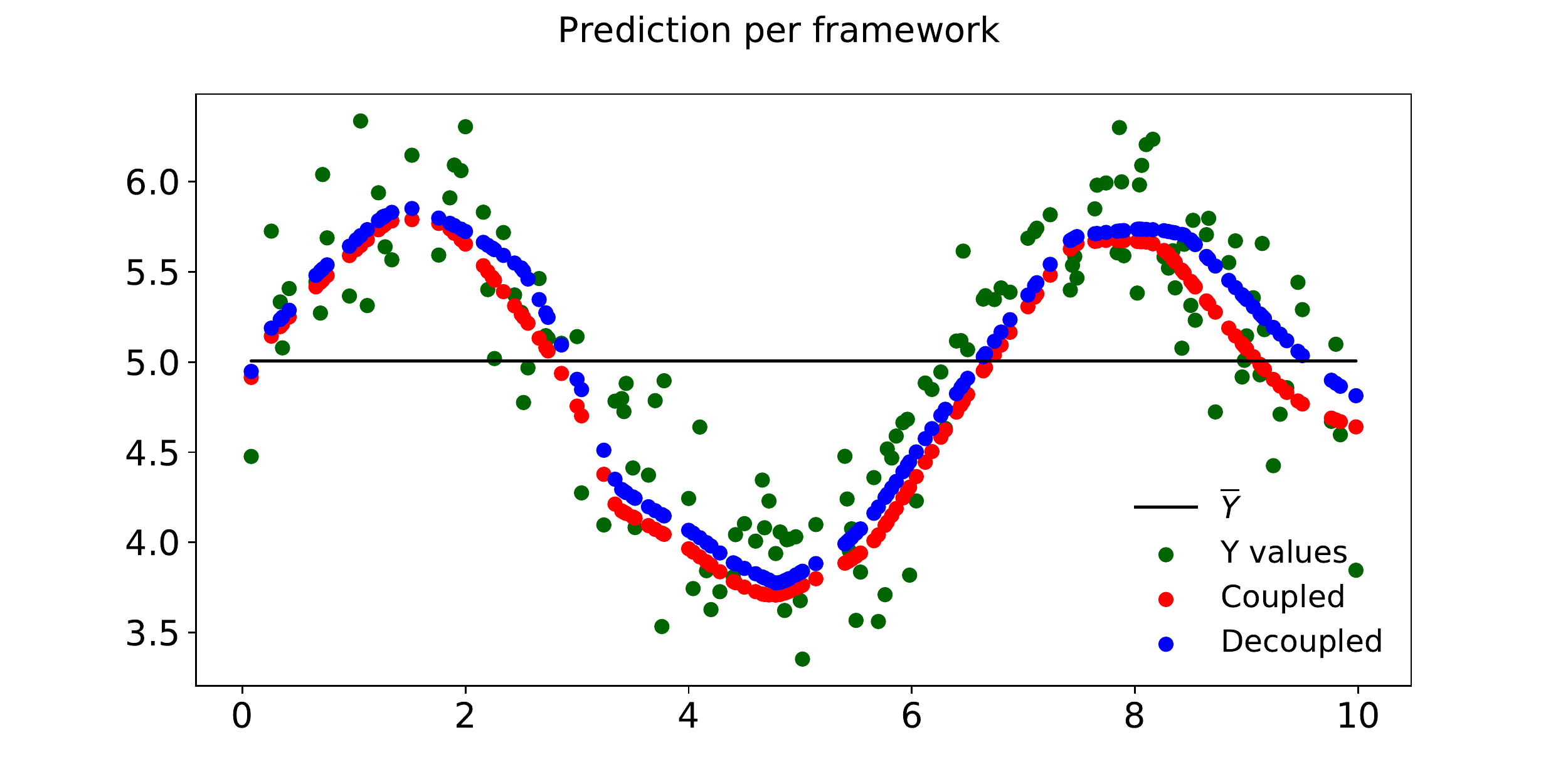}
        \caption{The NN model}
        
        \label{fig:sin_NN} 
    \end{subfigure}
    \caption{Generated data and ML predictions due to coupled and decoupled approaches for the sinusoidal sample}
    \label{fig:sin}
\end{figure}
\begin{figure}
    \centering
    \begin{subfigure}[h]{0.45\textwidth}
        \includegraphics[width=\textwidth]{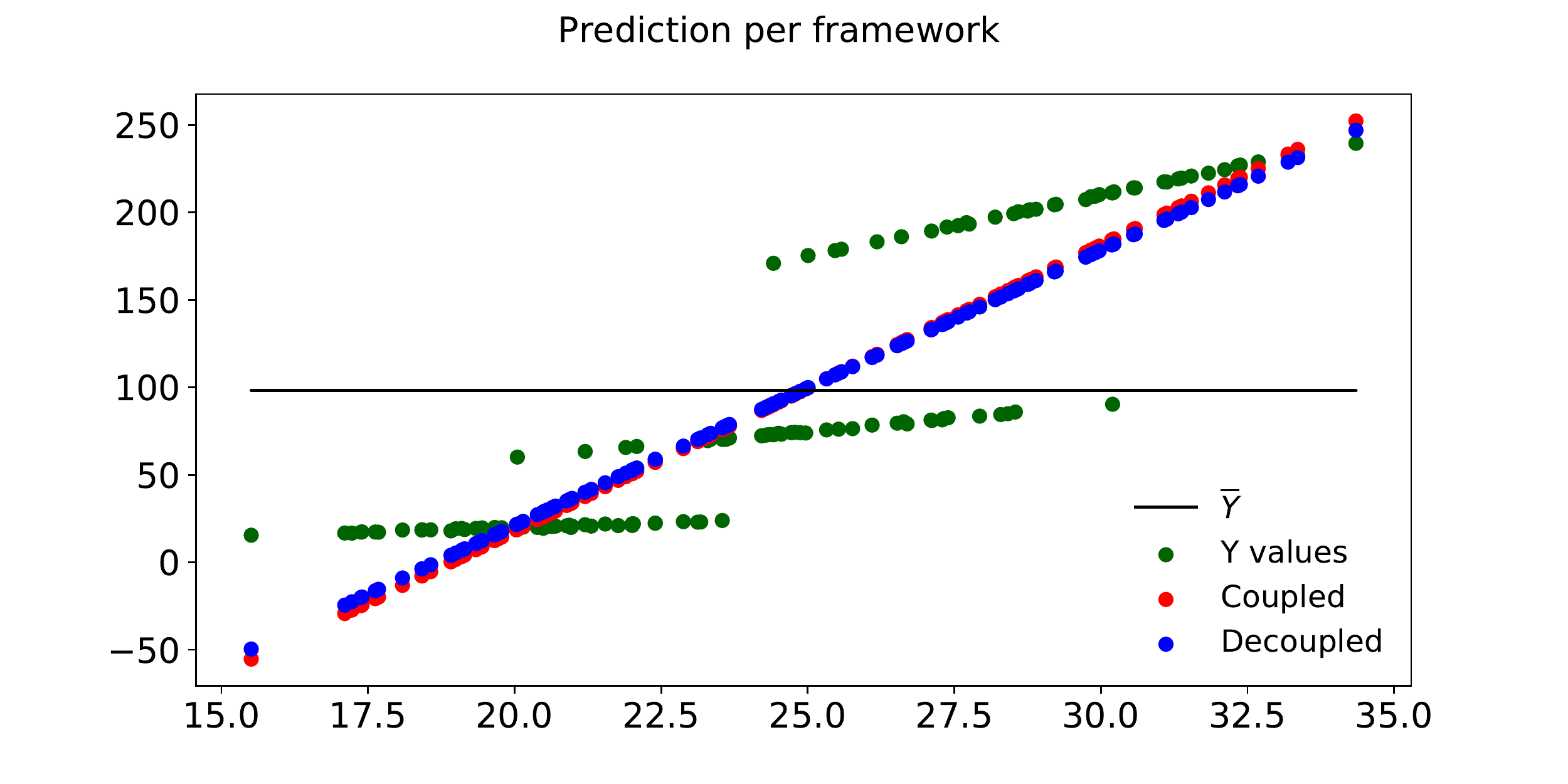}
        \caption{ The Lasso model}
        \label{fig:piece_lasso}       
    \end{subfigure}
    \begin{subfigure}[h]{0.45\textwidth}
        \includegraphics[width=\textwidth]{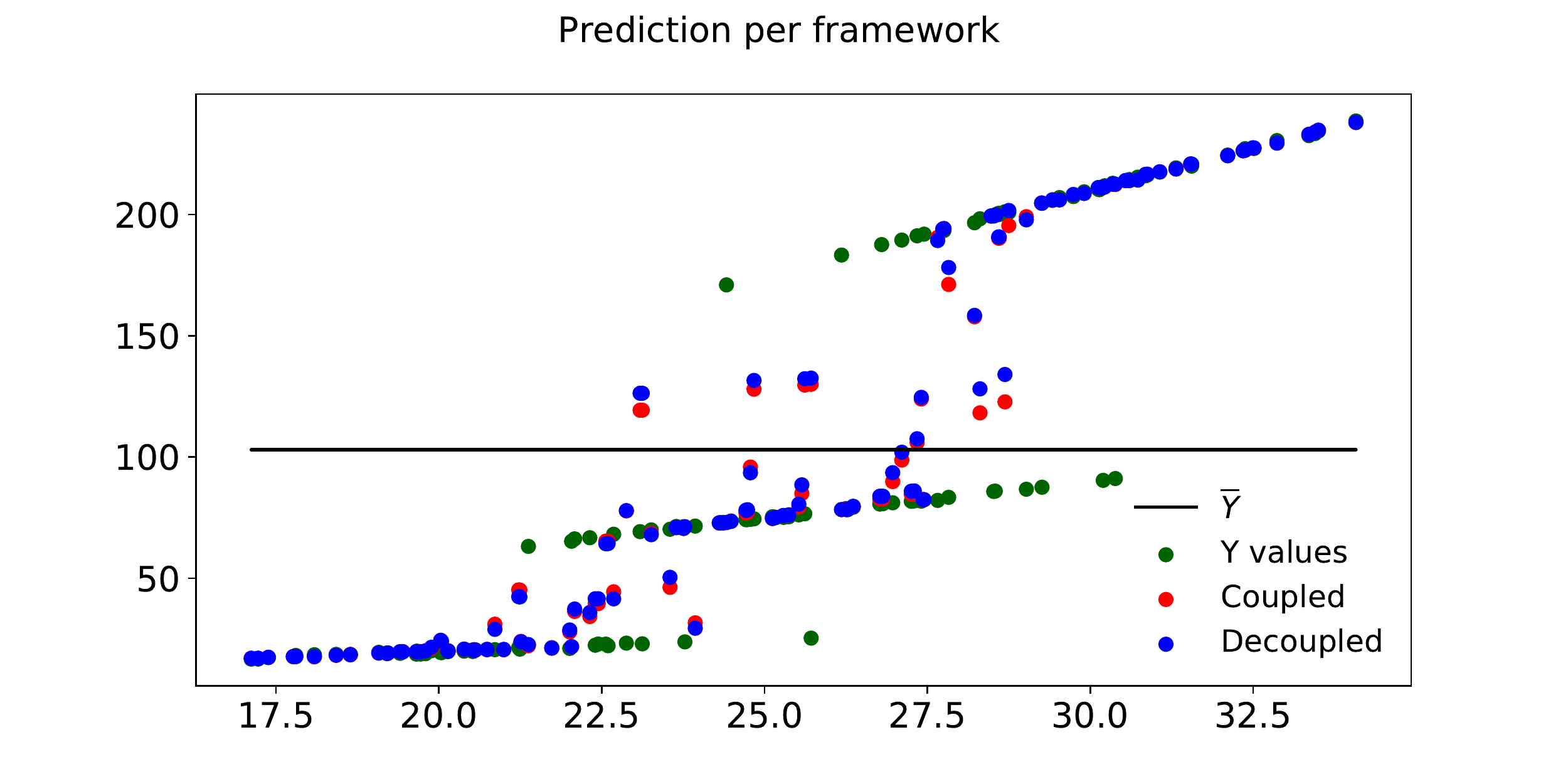}
        \caption{The RF model}
        \label{fig:piece_RF} 
    \end{subfigure}
        \begin{subfigure}[h]{0.45\textwidth}
        \includegraphics[width=\textwidth]{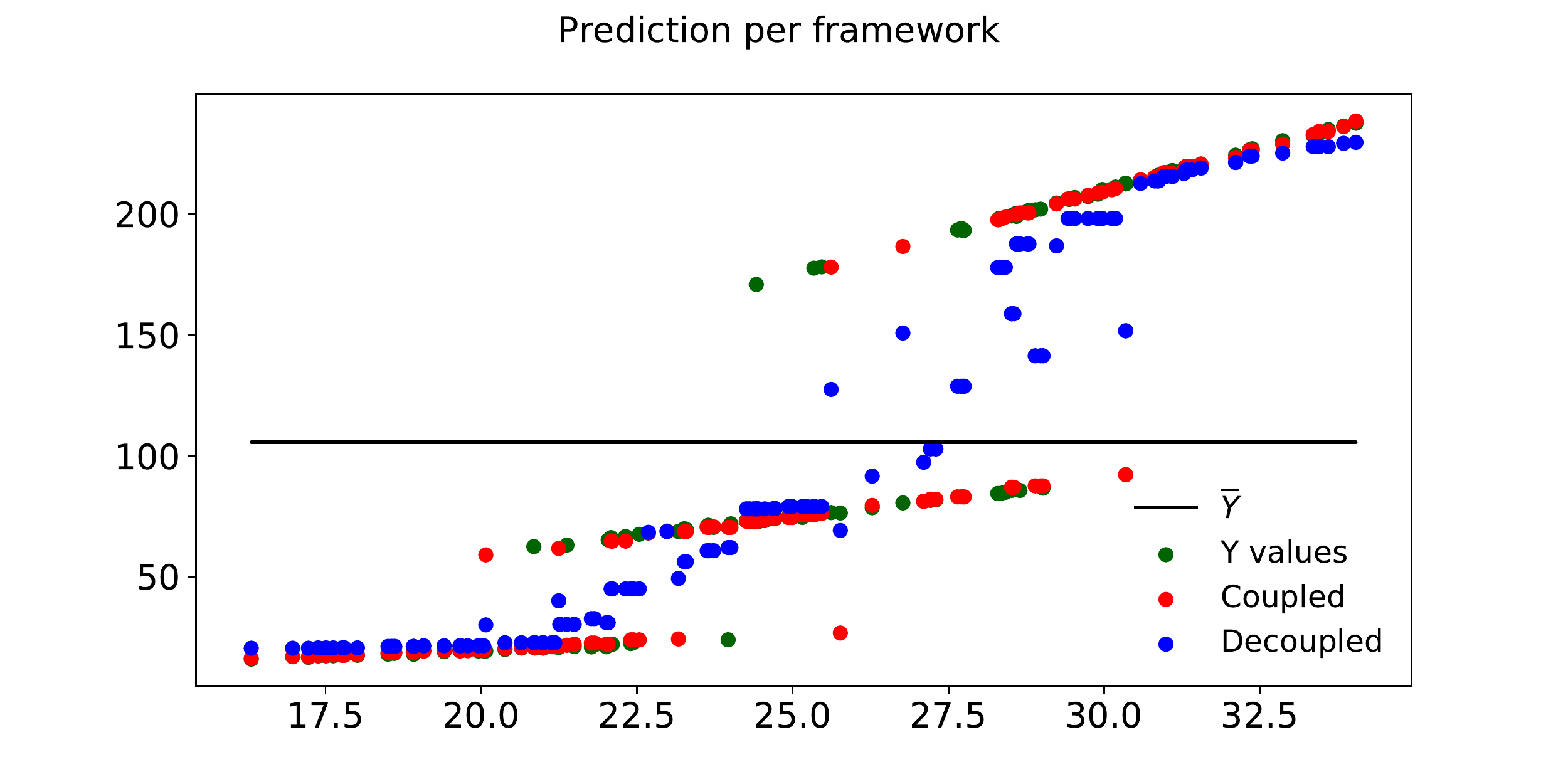}
        \caption{The GBM model}
        \label{fig:piece_GB} 
    \end{subfigure}
    \begin{subfigure}[h]{0.45\textwidth}
        \includegraphics[width=\textwidth]{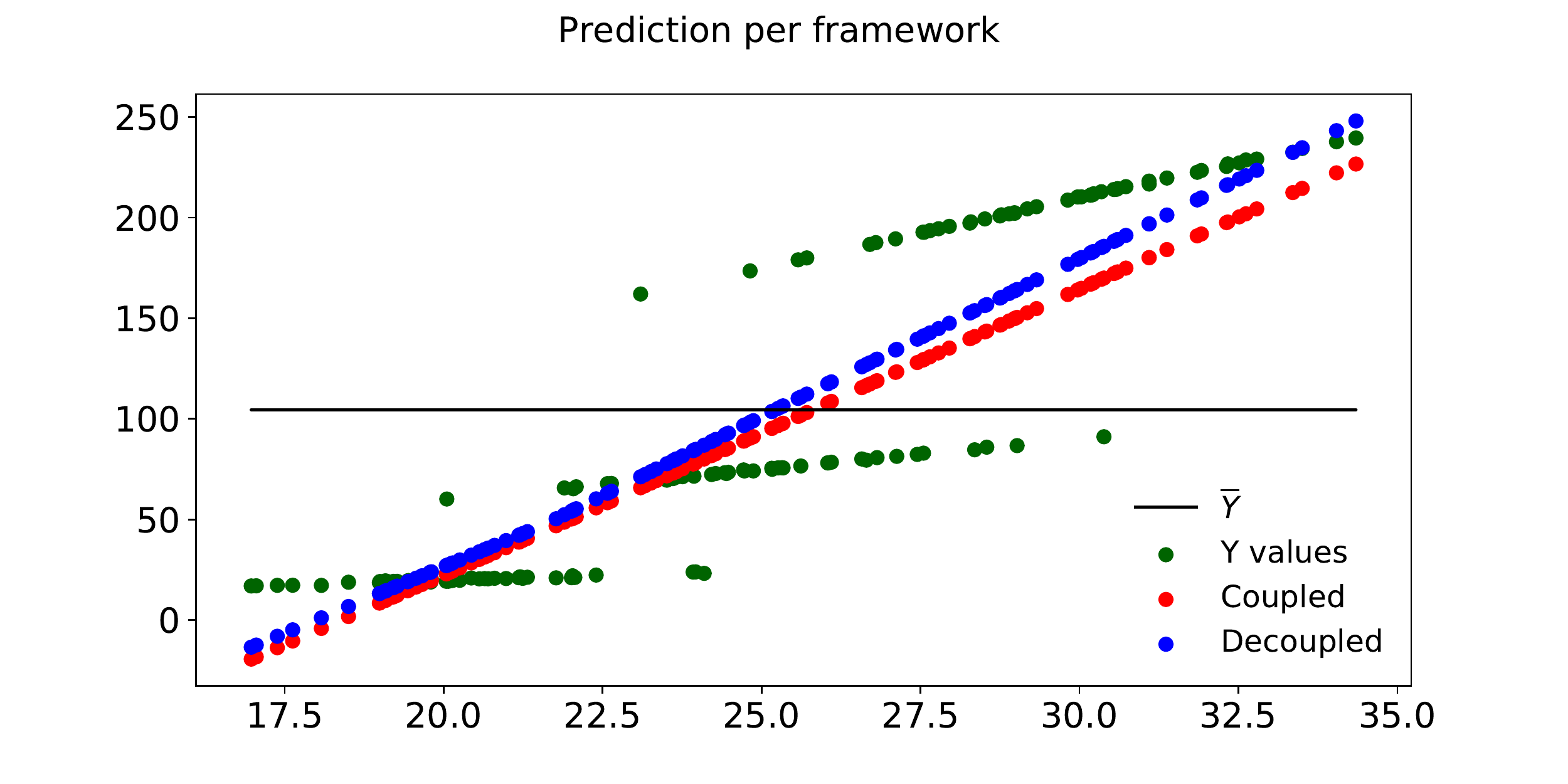}
        \caption{The NN model}
        \label{fig:piece_NN} 
    \end{subfigure}
    \caption{Generated data and ML predictions due to coupled and decoupled approaches for the piece-wise linear sample}
    \label{fig:piece}
\end{figure}

In the first experiment we set $n=1$, $a=1$ and
 use all four ML models to predict $\Y$. The generated values for both $x$ and $\Y$ are depicted in Fig~\ref{fig:identity_lasso}-~\ref{fig:piece_NN}. In the second set we set of experiments \mbox{$n = 10$}. Here though, we 
only employ neural network as a representative example, for the sake of reducing computational effort. 
We also add an intercept ($\ge1$) to the sinusoidal function to circumvent negative $\Y$ outcomes. 

The results for the first series ($n=1$) are summarized in Table~\ref{tab:deterministic}. 
First, consider the deterministic (pure prediction) models against the naive SAA. Observe that the former outperform the latter for the linear sample, and vice versa for sinusoidal and piece-wise linear samples. Naturally, since the linear relationship is simpler, the effect can be explained by higher prediction accuracy of all ML models in this case, illustrating the idea that  when the prediction accuracy is high enough, the deterministic models can be comparable to or even better than the stochastic models. Further, observe that in all cases the hybrid SAA-ML methods outperform both simple SAA and pure prediction.

Most importantly, observe that, as expected, the proposed coupled approach outperforms the decoupled one in almost all cases, both for deterministic and for hybrid SAA-ML models. The difference is not significant for some cases, but is more substantial in others, especially for the deterministic models. The difference is especially significant for neural network based models, which could be explained by observing that these tend to have more hyperparameters that should be selected in the validation step. 
Figures \ref{fig:identity}, \ref{fig:sin} and \ref{fig:piece} depict the relationship between the predictor and the simulated demand (in green), as well as the predicted values for all models (in red for coupled and in blue for decoupled) across the three samples. Observe that while in some cases the resulting predictions are very similar, in some (e.g., all neural network models), there exists a clear bias towards underestimating the demand, which is naturally beneficial to the prescriptive loss, as it leads to more conservative decisions.

In all cases, the best overall performance (other than the perfect information solution) is achieved by SAA-ML-Coupled approach, illustrating the power of the hybrid approach. For this method, the coupled validation leads to small but consistent improvement compared to the decoupled version.

Table~\ref{tab:10x} summarizes the obtained results for the case of multiple predictors ($n=10$). The observed performance of coupled vs decoupled approach is very similar to the simpler $n=1$ case.




\begin{table}[h!]
    \small
    \caption{Average prescription cost for solutions based on all methods and models for $n=1$ and the three samples. Values in bold represent the lowest cost (other than perfect information solution) for each sample}
	\centering
	\label{tab:deterministic}
\resizebox{0.70\textwidth}{!} {%
\begin{tabular}{llccc}
		
\hline
Method&Model& Linear sample &Sinusoidal sample &Piece-wise sample\\
\hline
\multirow{4}{*}{D-ML-Coupled}&Lasso&581.22&266.17&7734.21\\
&RF&499.11&154.39&6990.21\\
&GBM&498.23&160.14&6432.77\\
&NN&487.33&150.29&6752.31\\
\hline
\multirow{4}{*}{D-ML-Decoupled}&Lasso&582.33&267.99&7812.44\\
&RF&503.44&169.23&7089.34\\
&GBM&502.21&171.80&6499,32\\
&NN&488.12&188.24&6840.11\\
\hline
SAA&----&601.66&133.09&5330.92\\
SAA-ML-Coupled&\textit{k}NN&\textbf{421.76}&\textbf{111.73}&\textbf{4223.81}\\
SAA-ML-Decoupled&\textit{k}NN&438.91&113.01&4279.88\\
\hline
Perfect-info&----&390.22&99.91&2111.21\\
	\end{tabular}
}
\end{table}

\begin{table}[h!]
    \small
    \caption{Average prescription cost for solutions based on all methods and models for $n=10$ and the three samples. Values in bold represent the lowest cost (other than perfect information solution) for each sample }
	\centering
	\label{tab:10x}
\resizebox{0.9\textwidth}{!} {%
	\begin{tabular}{l|cc|c|cc|c}
		\hline
Sample& D-ML-Coupled &D-ML-Decoupled&SAA& SAA-ML-Coupled&SAA-ML-Decoupled&Perfect-info\\
\hline
Linear sample&4606.64&4703.82&4441.33&\textbf{4259.53}&4320.198&3999.04\\
Sinusoidal sample&166.74&192.33&145.19&\textbf{141.15}&143.48&98.33\\
Piece-wise sample&49395.22&49774.92&32921.44& \textbf{16144.98}&16199.36&15098.10\\

\hline
	\end{tabular}
}
\end{table}


\begin{table}[h!]
    \small
    \caption{Average prescription cost for solutions based on all methods for the real data. Values in bold represent the lowest cost (other than perfect information solution) for each sample }
	\centering
	\label{tab:real}
\resizebox{0.9\textwidth}{!} {%
	\begin{tabular}{l|cc|c|cc|c}
		\hline
Sample& D-ML-Coupled &D-ML-Decoupled&SAA& SAA-ML-Coupled&SAA-ML-Decoupled&Perfect-info\\
\hline
Concrete&2131.85&2155.85&1595.59&\textbf{1212.31}&1494.59&742.31\\

\hline
	\end{tabular}
}
\end{table}

\subsection{Real data}
To compare the coupled and decoupled approaches on a real-world dataset, we repurposed the standard Concrete Compressive Strength dataset given in~\cite{yeh} to be used in the context of the presented newsvendor problem. The dataset consists of 1030 records with 9 real-valued attributes with no missing values. The outcome, $\Y$ is assumed to represent demand, and the eight attributes are assumed to represent the auxiliary features impacting it. The results of solving the newsvendor problem in this case  are summarized in Table~\ref{tab:real}. Observe that in this case the obtained results are generally in agreement with the outcome of synthetic data experiments. Indeed, the proposed coupled approach outperforms the decoupled version both for pure prediction (deterministic) methods as well as hybrid SAA-ML version. Note though, that in this case the improvement is more significant for the latter comparison, further emphasizing the benefit of the proposed method.


\subsection{The impact of noise (model accuracy)}
\begin{figure}[h!]
\centering
        \includegraphics[width=0.75\textwidth]{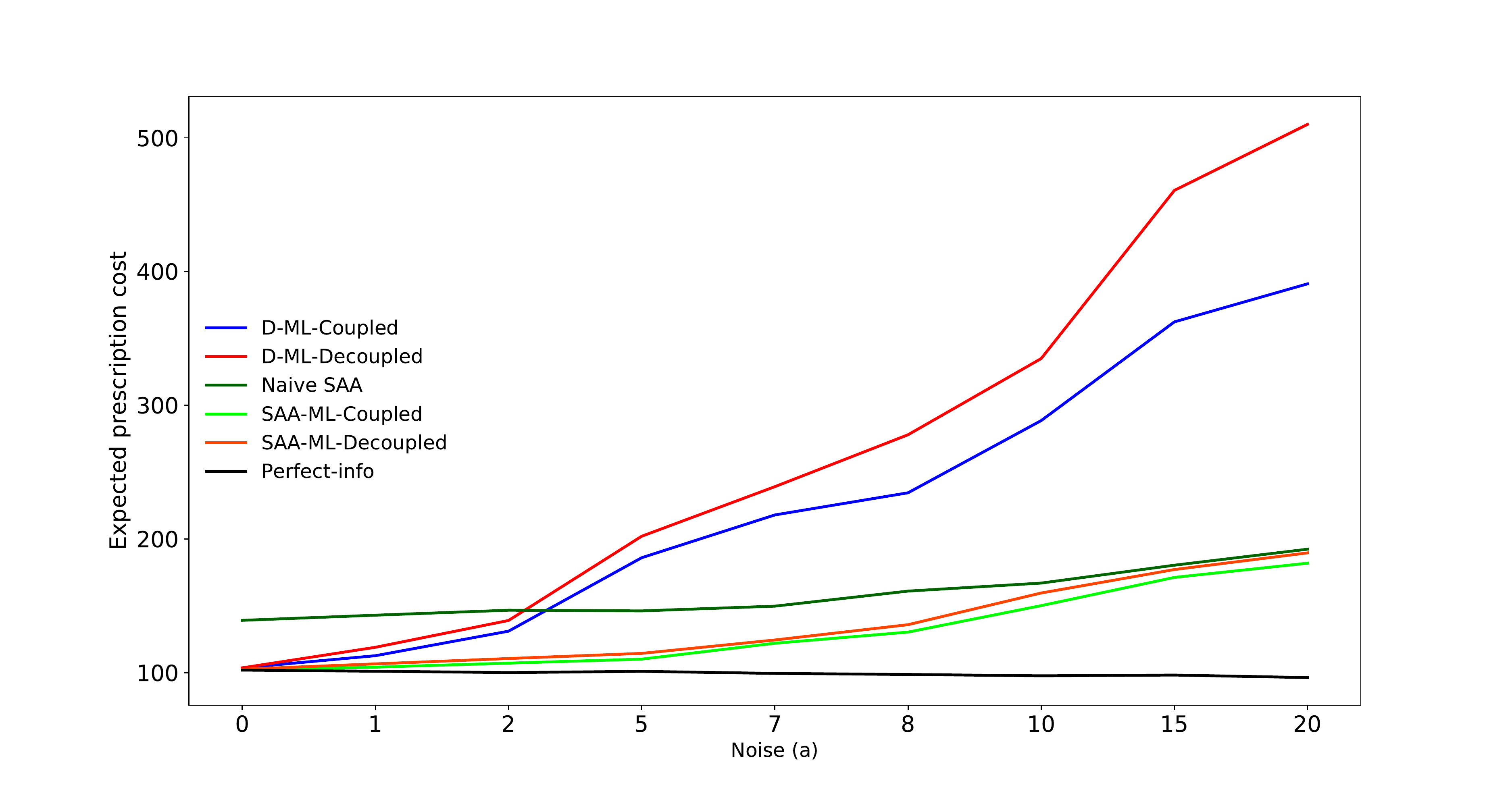}
        \caption{Expected prescription cost for the six models under sinusoidal sample with varied level of uniform noise.}
        \label{fig:noise}       
\end{figure}

To conclude the experiments, we evaluate the effect of the relative noise scale on the performance of the six decision making methods. To this end, we consider the synthetic data case with $n=1$ and sinusoidal function, and vary $a$ (uniform noise level) from 0 to 20. We again only consider the neural network version of the predictive model.    
Fig~\ref{fig:noise} presents the results of the experiment. Here, we plot the average  prescription cost for each of the 6 models against the noise level. Naturally, the perfect information solution is not dependent on the noise and the small discrepancy from the horizontal line can be explained with random resampling of the values themselves (in other words, if infinite samples where used, then the perfect information solution would result in a straight line). Naive SAA, on the other hand, naturally is dependent on the noise, since it relies on the historical records, which contain the noise factor. 

First, again observe that the coupled version of both deterministic ML model and hybrid SAA-ML outperforms the decoupled counterpart in all cases. The difference is relatively minor for SAA-ML method, but is significant for the D-ML models, especially for high levels of noise (i.e., when prediction accuracy is low). Secondly, the experiment confirms the expected behavior of the SAA vs deterministic models. If the prediction accuracy is high enough (noise is low), then the pure predictive model is superior to naive SAA (in this case there is no value in hedging against unfavorable outcomes, since those are not likely, given an accurate model). On the other hand, as soon as accuracy is reduced, the deterministic model performance quickly deteriorates and non-hedged decisions based on inaccurate prediction can lead to very high prescription cost. At the same time, in all levels of noise, the hybrid SAA-ML model outperforms both, i.e., it is always preferable to employ both predictive power of a (perhaps) an inaccurate ML prediction and stochastic optimization framework to protect against poor outcomes.

\newpage

\newpage

\section{Conclusions}
In this research effort we propose a new approach to training predictive models that are intended to be used in a prescriptive setting. Specifically, we argue that since in this case  prediction accuracy is not the ultimate goal of training, the prescription cost (or reward) should be explicitly incorporated in ML model development. We propose to employ the validation step for this purpose and consider an approach, referred to as coupled validation, which achieves this goal by defining prescription cost as the objective in the hyperparameter selection step of the standard machine learning framework. We demonstrate that it posses promising theoretical properties  relative to the standard (decoupled) method. 
Further, we revisit recently proposed hybrid prediction-stochastic-optimization framework due to \cite{Bertsimas}, and demonstrate that our approach is also applicable in this case.

We perform a series of experiments illustrating the extent to which the theoretical benefit of the  coupled approach translates to the realized improvement in both synthetic and real datasets applied to the classic newsvendor problem. Specifically, we observe that while it outperforms decoupled version in almost all cases, the improvement can vary from relatively small to substantial.  We also evaluate the effect of model noise (and hence model accuracy) on the method performance. 

We argue that the proposed method possesses a number of benefits. It is easy to implement since it naturally fits into the standard training-validating-testing scheme. Further, it is general and predictive mode independent, and hence, can be applied in a model-agnostic fashion. On the other hand, there are a number of points worth heeding. First, our experiments showed that the required computational effort for training mostly depends on the underlying optimization problem. That is, we expect that the coupled approach to require more significant computational effort than the decoupled approach when the optimization problem is more challenging to solve (e.g. NP-hard class of optimization problems~\cite{hochba}). Second, the coupled approach introduces bias in predictions by design. Therefore, the trained predictive models with the coupled approach are not suitable to be used in pure prediction tasks.

\bibliography{ref}


\section*{Appendix}
\begin{table}[h!]
    \small
    \begin{center}

	\caption{Training, validating and testing steps for the coupled approach in SAA-ML model based on  $k$NN}
	\label{tab:SAA-ML-C}
\resizebox{.55\textwidth}{!} {%
	\begin{tabular}{l}
\hline
   neighbors $\leftarrow$ \textbf{len}($\x^r$) \\
     \textbf{for} $n$ in \textbf{range}\:(1,\: neighbors):\\
 \qquad         SAA\_ML\_coupled $\leftarrow [\:]$\\
 \qquad         \textbf{for} each $\x^v$:\\
 \qquad \qquad            distance $\leftarrow\{\:\}$\\
 \qquad \qquad             \textbf{for} $j$ in $\x^r$:\\
 \qquad \qquad \qquad           distance$[j] \leftarrow \|\x^v-\x^r_j\|^2$\\
 \qquad  \qquad           distance\_sorted $\leftarrow$ \textbf{sort}(\mbox{distance.values(ascending=true)})\\
 \qquad \qquad           index $\leftarrow$ \mbox{distance\_sorted[1:n])}\\
  \qquad \qquad       $\z \leftarrow \mbox{saa}(\y^r$[index])\\
 \qquad  \qquad           SAA\_ML\_coupled.\textbf{append}($g(\z, \y^v)$)\\
 \qquad       SAA\_ML\_coupled\_means$[n] \leftarrow \textbf{mean}(\mbox{SAA\_ML\_coupled)}$\\
  \\
  best\_n $\leftarrow \argmin_n \mbox{(SAA\_ML\_coupled\_means)}$\\
  \\
   SAA\_ML\_coupled$\leftarrow$ [\:]\\
   \textbf{for} $\x^t$:\\
 \qquad      distance $\leftarrow$ \{\}\\
 \qquad     \textbf{for} $j$ in ($\x^{rv})$:\\
 \qquad \quad            distance[j] $\leftarrow \|\x^t-\x^{rv}_j\|^2$\\
 \qquad             distance\_sorted $\leftarrow$ \textbf{sort}(\mbox{distance.values(ascending=true))}\\
 \qquad           index $\leftarrow$ \mbox{distance\_sorted[1:best\_n])}\\
 \qquad                  $\z \leftarrow$ \mbox{saa($y^{rv}$[index]}\\
 \qquad       SAA\_ML\_coupled\_temp.\textbf{append}($g(\z, \y^t)$)\\
 SAA\_ML\_coupled\_final $\leftarrow$ \textbf{mean}(\mbox{SAA\_ML\_coupled\_temp)}\\
 \textbf{return} SAA\_ML\_coupled\_final\\
\hline
	\end{tabular}}
	\end{center}
\end{table}

\begin{table}[h!]
    \small
\begin{center}
    
\caption{Training, validating and testing steps for the decoupled approach in SAA-ML model based on $k$NN}

\label{tab:SAA-ML_D}
\resizebox{.55\textwidth}{!}{%
	\begin{tabular}{l}
\hline
  neighbors $\leftarrow$ \textbf{len}($\x^r$) \\
    \textbf{for} $n$ in \textbf{range}\:(1,\: neighbors):\\
\qquad         SAA\_ML\_decoupled$\leftarrow [\:]$\\
\qquad         \textbf{for}$\x^v$:\\
\qquad \qquad            distance$\leftarrow$\{\:\}\\
\qquad \qquad             \textbf{for} $j$ in $\x^r$:\\
\qquad \qquad \qquad           distance$[j]\leftarrow \|\x^v-\x^r_j\|^2$\\
\qquad  \qquad           distance\_sorted$\leftarrow$ \textbf{sort}(\mbox{distance.values(ascending=true)})\\
\qquad \qquad           index$\leftarrow$ \mbox{distance\_sorted[1:n]}\\
\qquad  \qquad           SAA\_ML\_decoupled.\textbf{append}(\textbf{mean}($\y^r$[index]))\\
\qquad      SAA\_ML\_decoupled\_means[n]$\leftarrow$\textbf{mean}(\mbox{SAA\_ML\_decoupled)}\\
 \\
  best\_n$\leftarrow$ $\argmin_n\mbox{(SAA\_ML\_decoupled\_means)}$\\
 \\
  SAA\_ML\_decoupled$\leftarrow[\:]$\\
  \textbf{for} $\x^t$:\\
\qquad      distance $\leftarrow$ \{\}\\
\qquad     \textbf{for} $j$ in $\x^{rv}$:\\
\qquad \quad            distance$[j]\leftarrow\|\x^t-\x^{rv}_j\|^2$\\
\qquad             distance\_sorted$\leftarrow$\textbf{sort}\mbox{(distance.values(ascending=true))}\\
\qquad           index$\leftarrow$\mbox{distance\_sorted[1:best\_n])}\\
\qquad               \z$\leftarrow$\mbox{saa($y^{rv}$[index])} \\
\qquad       SAA\_ML\_decoupled\_temp.\textbf{append}($g(\z, \y^t)$)\\
SAA\_ML\_decoupled\_final$\leftarrow$ \textbf{mean}(\mbox{SAA\_ML\_decoupled\_temp)}\\
\textbf{return} SAA\_ML\_decoupled\_final\\
\hline
	\end{tabular}}
\end{center}
\end{table}

\end{document}